\documentclass[runningheads]{llncs}

\usepackage[mobile]{eccv}

\usepackage{eccvabbrv}

\usepackage{graphicx}
\usepackage{booktabs}
\usepackage[nointegrals]{wasysym}
\usepackage{pifont}
\usepackage{multirow}
\usepackage{amssymb}
\usepackage{makecell}
\usepackage{authblk}

\usepackage[accsupp]{axessibility}
\usepackage{orcidlink}

\usepackage{hyperref}

\begin{document}

\title{Robust 3DGS-based SLAM via \\Adaptive Kernel Smoothing} 



\author{
Shouhe Zhang\inst{1} \and
Dayong Ren\inst{2}\textsuperscript{*} \and
WEN JIE LI\inst{3} \and
Piaopiao Yu\inst{4} \and
Sensen Song\inst{1}\textsuperscript{*} \and
Kaikai Shao\inst{5} \and
Yurong Qian\inst{1}
}

\authorrunning{M.Zhang et al.}

\institute{School of Computer Science and Technology, Xinjiang University \and The National Key Laboratory for Novel Software Technology, Nanjing University
\and Southwest University of Political Science and Law
\and Nanjing University of Aeronautics and Astronautics
\and School of Physics and Electronic Information, Yantai University
}

\maketitle

\begingroup
\renewcommand{\thefootnote}{\fnsymbol{footnote}}
\footnotetext[1]{Corresponding author.Email:rdyedu@gmail.com. songsensen@xju.edu.cn
}
\endgroup

\begin{abstract}
In this paper, we challenge the conventional notion in 3DGS-SLAM that rendering quality is the primary determinant of tracking accuracy. We argue that, compared to solely pursuing a perfect scene representation, it is more critical to enhance the robustness of the rasterization process against parameter errors to ensure stable camera pose tracking. To address this challenge, we propose a novel approach that leverages a smooth kernel strategy to enhance the robustness of 3DGS-based SLAM. Unlike conventional methods that focus solely on minimizing rendering error, our core insight is to make the rasterization process more resilient to imperfections in the 3DGS parameters. We hypothesize that by allowing each Gaussian to influence a smoother, wider distribution of pixels during rendering, we can mitigate the detrimental effects of parameter noise from outlier Gaussians. This approach intentionally introduces a controlled blur to the rendered image, which acts as a regularization term, stabilizing the subsequent pose optimization. While a complete redesign of the rasterization pipeline is an ideal solution, we propose a practical and effective alternative that is readily integrated into existing 3DGS frameworks. Our method, termed Corrective Blurry KNN (CB-KNN), adaptively modifies the RGB values and locations of the K-nearest neighboring Gaussians within a local region. This dynamic adjustment generates a smoother local rendering, reducing the impact of erroneous GS parameters on the overall image. Experimental results demonstrate that our approach, while maintaining the overall quality of the scene reconstruction (mapping), significantly improves the robustness and accuracy of camera pose tracking.\url{https://github.com/xju-zsh/Robust-3DGS-based-SLAM-via-Adaptive-Kernel-Smoothing.git}
  \keywords{3DGS, SLAM, Smooth Kernel, Corrective Blurry KNN}
\end{abstract}

\section{Introduction}
\label{sec:intro}

Simultaneous Localization and Mapping (SLAM) is a cornerstone of robotics and computer vision, enabling autonomous agents to navigate unknown environments while concurrently constructing a map\cite{3dgs,33,47}. The field has seen remarkable advancements, particularly with the advent of real-time dense reconstruction methods\cite{MonoGS,37,49,54}. Among these, the recent introduction of 3D Gaussian Splatting (3DGS) has garnered significant attention for its ability to render high-fidelity neural radiance fields at unprecedented speeds\cite{Gs-slam,nerf,34,35}. This paradigm shift from implicit neural representations to an explicit, point-based structure has opened new avenues for integrating rich scene representations directly into the SLAM loop\cite{plenoctrees}. Many modern 3DGS-based SLAM systems leverage this fast rasterization process, using the rendered depth and color images for photometric and geometric alignment to estimate camera pose\cite{neural,rpbg,36,39}.

Despite the promising real-time performance, we have identified a critical vulnerability in current 3DGS-based SLAM systems. The core challenge lies in the inherent inaccuracies of the high-dimensional Gaussian parameters\cite{Orb-slam3,direct-sparse-odometry,enhanced-LiDAR-Based}. These parameters, which include position, orientation, scale, and color, are optimized from noisy camera observations\cite{LSD-SLAM,elasticfusion,41}. Consequently, the peripheral regions of the Gaussian kernels often contain significant errors. During the rasterization process, these erroneous parameters can lead to spurious artifacts in the rendered image, manifesting as incorrect pixel colors or even discontinuous holes in the scene\cite{large-scale-photometri,online-photometric}. These artifacts introduce substantial noise into the photometric alignment used for pose tracking, directly compromising the accuracy and robustness of the camera's localization, particularly in challenging conditions like fast motion or aggressive viewpoint changes\cite{TUM-RGBD,replica-dataset,38,45,67}.

\begin{figure}[t!]
\centering
\includegraphics[scale=0.32, trim=0cm 0cm 0cm 0cm, clip]{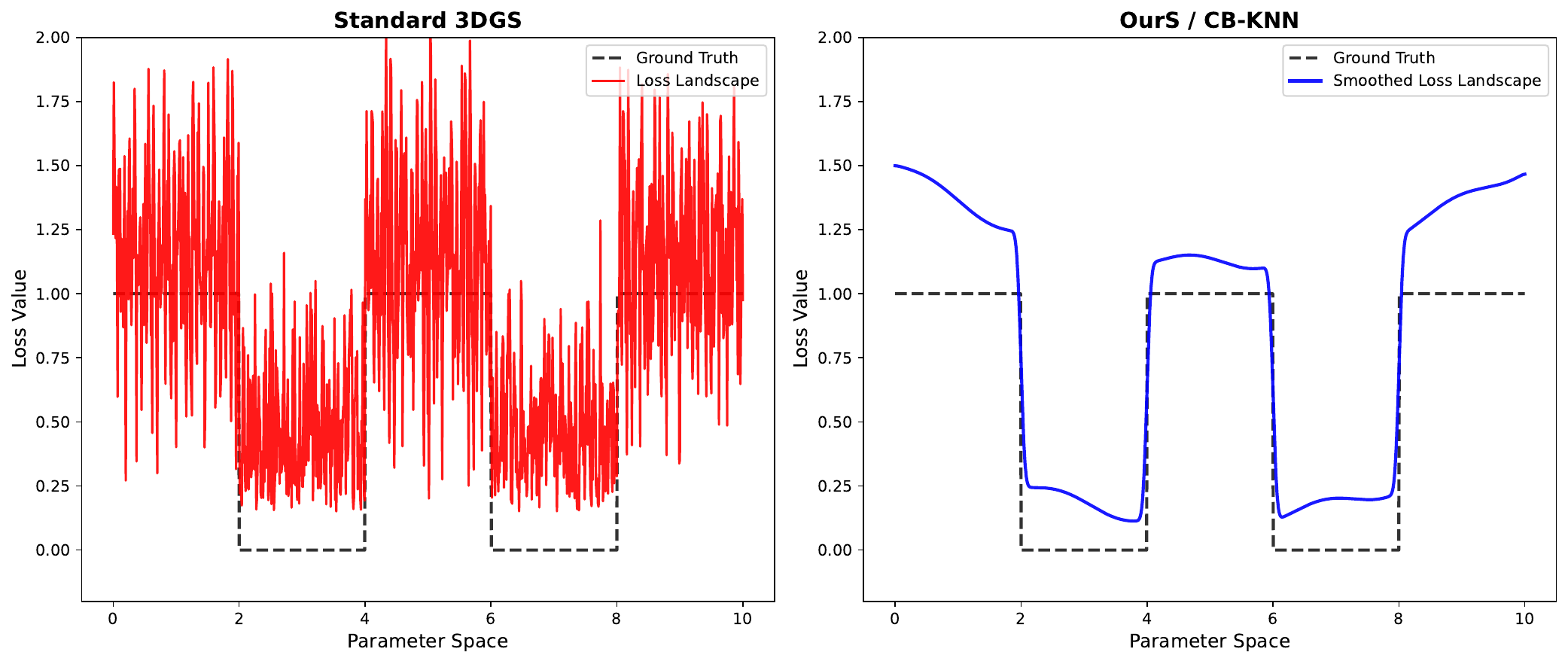}
\caption{Optimization Landscape Analysis. High-frequency noise in 3DGS
parameters often creates a rugged non-convex loss landscape (Left), causing
gradient-based optimization to get stuck in sharp local minima. Our CB-KNN
strategy acts as a run-time regularization (Right), effectively smoothing the
landscape and expanding the basin of attraction, ensuring robust
convergence even under significant parameter noise.}
\label{fig:loss_landscape_comparison}
\end{figure}

Traditional SLAM paradigms often operate under the assumption that a high-fidelity, visually sharp representation is essential for accurate pose estimation\cite{euroc,66}. However, in the context of 3DGS, we argue that this may not always be true. Our core insight is that for the purpose of pose tracking, a slightly smoother or blurred rendered image can be more resilient to the noise inherent in the Gaussian parameters\cite{41,42,58}.Theoretically, this strategy aligns with the principles of Graduated Non-Convexity (GNC). Photometric alignment in 3DGS-SLAM is a highly non-convex problem where rendering artifacts—caused by noisy Gaussian centers or scales—act as "sharp" local minima that trap the optimizer. By enabling each Gaussian to influence a wider, more continuous region of pixels, we effectively perform a smoothing operation on the cost function's landscape(as conceptually illustrated by the 1D toy example in Fig. \ref{fig:loss_landscape_comparison}). This allows the system to bypass the deleterious local minima induced by outlier Gaussians, facilitating a more stable and global convergence for camera pose tracking.This controlled smoothing effect dampens the potentially disruptive impact of individual outlier Gaussians, producing a more stable and reliable image for subsequent pose optimization, thereby decoupling the often conflicting requirements for scene fidelity and tracking robustness\cite{43,48,68}.

To achieve adaptive kernel smoothing while remaining compatible with existing 3DGS frameworks, we propose Corrective Blurry KNN (CB-KNN)—a practical, plug-and-play correction mechanism. Unlike a complex and rigid full redesign of the rasterization pipeline\cite{applications-of-Robust}, CB-KNN seamlessly integrates into existing workflows by introducing a local neighborhood-based smoothing term during pixel rendering. Specifically, for each pixel, the rasterizer identifies the K-nearest neighboring Gaussians and dynamically adjusts their parameters\cite{online-photometric,applications-of-Robust}: first, it shifts their locations toward their collective centroid to enhance 2D overlap and fill structural gaps\cite{large-scale-photometri}; second, it computes a locally adaptive weighted average of their RGB values to suppress the outlier-induced color noise and improve rendering robustness\cite{3dgs}.

Crucially, these corrections are transient and preserve the underlying 3DGS parameters\cite{3dgs,42,44,51}. This temporary regularization enhances pose tracking robustness without compromising scene integrity\cite{MonoGS,Gs-slam}. To mitigate the computational overhead of large-scale SLAM, CB-KNN is applied exclusively to keyframes during mapping\cite{plenoctrees}. Since keyframes anchor map quality and pose accuracy\cite{rpbg}, stabilizing their rendering provides a robust foundation for subsequent tracking\cite{physically-Based}—achieving superior accuracy while remaining computationally viable for mobile scenarios.

The key contributions of this work are as follows:

\begin{enumerate}

\item We introduce a novel perspective for 3DGS-based SLAM, shifting the focus from maximizing visual fidelity to ensuring the robustness of pose estimation.

\item We propose the concept of adaptive kernel smoothing to mitigate the detrimental effects of parameter noise on the rasterized image.

\item We present Corrective Blurry KNN (CB-KNN), a practical and effective method that can be seamlessly integrated into existing 3DGS frameworks.

\item We provide comprehensive experimental results demonstrating that our approach significantly improves tracking accuracy and robustness while preserving mapping quality.

\end{enumerate}

\section{Related Work}

\subsection{3D Gaussian Splatting in SLAM}
While 3D Gaussian Splatting (3DGS) \cite{3dgs} has revolutionized real-time rendering by overcoming the bottlenecks of implicit representations like NeRF \cite{nerf}, its integration into SLAM systems \cite{Gs-slam,Co-slam,Gaussian-SLAM,Eslam,36,37} reveals a critical limitation. Existing 3DGS-based SLAM frameworks predominantly prioritize rendering clarity and scene fidelity. Consequently, they remain vulnerable to Gaussian parameter errors. During rapid camera movements or drastic viewpoint changes \cite{Vox-fusion}, these inaccuracies manifest as rendering artifacts, leading to photometric alignment failures and severe pose drift \cite{Gaussian-SLAM}. This highlights an urgent need to decouple scene fidelity from tracking robustness by rethinking 3DGS rendering strategies.

\subsection{Limitations of Traditional Robustness Strategies}
Conventional SLAM systems, such as ORB-SLAM3 \cite{Orb-slam3}, typically achieve robustness by dynamically weighting feature points or employing robust loss functions to mitigate observation noise\cite{47}. However, these traditional post-processing strategies are ineffective for 3DGS-based SLAM, where noise intrinsically originates from map representation errors (i.e., flawed Gaussian parameters) rather than mere sensor noise \cite{3dgs,Gaussian-SLAM}. Addressing this issue requires active regularization directly within the rendering pipeline, which fundamentally distinguishes our approach from conventional methods.

\subsection{Kernel Smoothing for 3DGS Rendering}
Kernel smoothing is a well-established technique for data regularization via local weighted averaging \cite{LiDAR-smoothing-and-mapping-with-planes,52}. Yet, its prior applications in SLAM have been restricted to 2D image denoising \cite{Adaptive-kerne,56}, 3D point cloud processing \cite{point-cloud,63,64,65,61,69}, or data association \cite{LiDAR-smoothing-and-mapping-with-planes}, leaving a gap in the 3DGS rendering process. Our proposed CB-KNN method pioneers the application of kernel smoothing in 3DGS by introducing three key innovations: (1) dynamic, dual-dimensional correction of both spatial positions and RGB values; (2) the utilization of kernel smoothing as a transient rendering regularizer that preserves the underlying canonical map; and (3) a selective application strategy focused exclusively on keyframes to ensure computational efficiency.

\section{Method}

\begin{figure}[t!]
\centering
\includegraphics[scale=0.14, trim=0cm 0cm 0cm 0cm, clip]{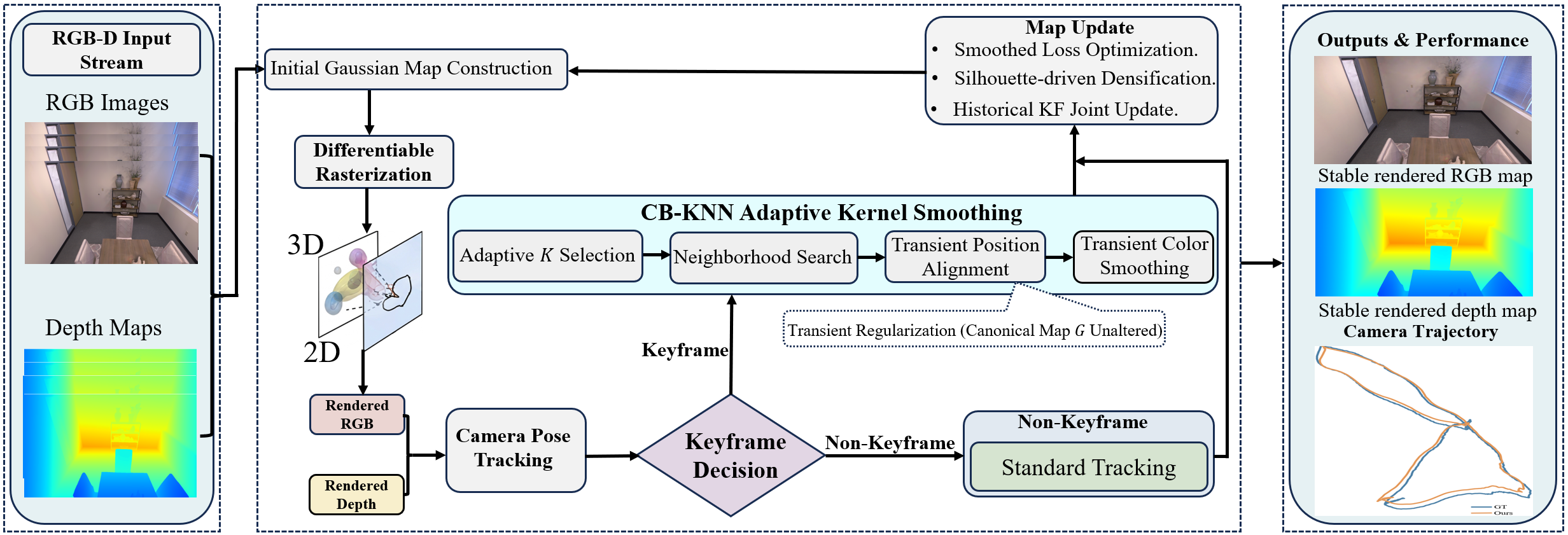}
\caption{Overview of the Robust 3DGS-SLAM Framework. The system processes synchronous RGB-D streams through the core Corrective Blurry KNN (CB-KNN) adaptive kernel smoothing mechanism. During the keyframe stage, transient regularization is introduced by adaptively selecting the neighborhood size $K$ and dynamically correcting Gaussian positions and colors. This process constructs a smoothed optimization landscape, effectively mitigating the impact of parameter noise to ensure robust camera tracking and high-fidelity scene reconstruction.}
\label{fig:framework}
\end{figure}

In this section, we present the systematic integration of 3D Gaussian Splatting (3DGS) \cite{3dgs} with a robust SLAM framework, centered around our proposed Corrective Blurry K-Nearest Neighbors (CB-KNN) adaptive kernel smoothing strategy. Our methodology is anchored by three foundational principles: transient rendering correction, keyframe-based smoothing, and preservation of original map fidelity. This design effectively suppresses the detrimental effects of Gaussian parameter noise during the differentiable rendering process, while ensuring seamless compatibility with end-to-end pose optimization and map refinement. The comprehensive workflow of the proposed CB-KNN-based SLAM system is illustrated in Fig. \ref{fig:framework}.

\subsection{Gaussian Representation and Differentiable Rendering.}
We employ a collection of 3D Gaussians as an explicit scene representation. To enhance the robustness of the tracking process, we integrate the CB-KNN correction mechanism into the rendering pipeline while strictly preserving the differentiability of 3DGS to support end-to-end pose and map optimization. This subsection details our approach across three dimensions: the definition of Gaussian parameters, the design of the rendering pipeline, and the generation of smoothed image modalities.

\textbf{3D Gaussian Map Representation.} The scene is represented by a set of 3D Gaussians ${G} = \{g_1, g_2, \dots, g_N\}$. Each individual Gaussian $g_i$ is defined by a tuple $(\mu_i, \sigma_i, r_i, c_i)$, where $\mu_i \in \mathbb{R}^3$ denotes the world-space center position, $\sigma_i \in [0, 1]$ represents the opacity, $r_i$ is the Gaussian radius, and $c_i$ specifies the view-independent RGB color. We adopt view-independent colors to ensure the isotropy of the Gaussian distribution, facilitating more stable optimization. Following \cite{3dgs}, the spatial influence of a Gaussian is characterized by a radial decay function:
\begin{equation}
\label{eq:Formula 1}
f(x)=\sigma_i\cdot\exp\left(-\frac{\|x - \mu_i\|^2}{r_i^2}\right)
.\end{equation}

Crucially, the parameters $(\mu_i, \sigma_i, r_i, c_i)$ stored in the map are treated as the canonical parameters. The subsequent CB-KNN corrections are applied exclusively to transient parameters during the rendering phase, ensuring that the underlying global map remains unaltered and consistent.

\textbf{CB-KNN-Enhanced Differentiable Rendering.}
Our framework integrates the CB-KNN correction directly into the differentiable rendering pipeline at the CUDA level, enabling robust synthesis of color, depth, and silhouette images. The process begins with standard frustum culling and depth-based sorting to identify visible Gaussians. For each pixel $p = (u, v)$, we adaptively select a local neighborhood ${G}_p = \{g_{p1}, \dots, g_{pK}\}$ consisting of the $K$ nearest Gaussians that contribute most significantly to the pixel value.

Within this local cluster, we perform transient corrections: Gaussian positions $(\mu_{p_k})$ and colors $(c_{p_k})$ are dynamically adjusted to generate a rectified set ${G}'_p$. This adaptive smoothing effectively bridges gaps and suppresses color noise caused by parameter inaccuracies. Crucially, these corrections are frame-specific and transient---they regularize the current rendering pass to facilitate stable pose optimization without permanently altering the underlying canonical map ${G}$.

\subsection{Formulation of CB-KNN Regularized Rendering}
To mathematically formalize the CB-KNN correction and its integration into the rasterization process, we detail the dual-dimensional smoothing calculation followed by the regularized image synthesis. This controllable smoothing acts as an on-the-fly regularization term, mitigating the impact of erroneous Gaussians.

\textbf{Adaptive Parameter Correction.} For each pixel $p = (u, v)$, let ${G}_p$ denote the set of its $K$-nearest contributing Gaussians. We extend the K-nearest neighbor logic to concurrently correct both the spatial locations and colors of the Gaussians in ${G}_p$. For each Gaussian $g_{pk} \in {G}_p$, we first compute the 2D projected centroid $C_p$ of the local neighborhood:
\begin
{equation}C_p=\frac{1}{K}\sum_{g_{pj}\in G_p}\pi(\mu_{pj},E_t,K_{cam})
.\end{equation}
where $\pi(\cdot)$ denotes the 3D-to-2D projection function parameterized by the camera extrinsics $E_t$ and intrinsics $K_{cam}$. To enforce spatial cohesion and bridge artifactual gaps, the 2D projection of $g_{pk}$ is subtly shifted towards centroid $C_p$:
\begin
{equation}\pi(\mu_{pk}^{\prime})=\pi(\mu_{pk})+\alpha\cdot\frac{C_p-\pi(\mu_{pk})}{\|C_p-\pi(\mu_{pk})\|+\epsilon}
.\end{equation}
Here, $\alpha \in [0.1, 0.3]$ controls the shift magnitude, $\epsilon = 10^{-6}$ prevents division by zero, and $\mu'_{p_k}$ represents the transiently corrected 3D position. Concurrently, to suppress high-frequency color noise, we assign normalized contribution weights $\omega_{p_j}$ to compute the smoothed color $c'_{p_k}$:
\begin{equation}c_{pk}^{\prime}=\sum_{g_{pj}\in G_p}\omega_{pj}\cdot c_{pj},\quad\omega_{pj}=\frac{f_{pj}(p)}{\sum_{g_{pj}\in G_p}f_{pj}(p)}.\end{equation}
This ensures that morphologically dominant Gaussians dictate the local color consensus.

\textbf{Regularized Image Synthesis.} Having obtained the transient parameter set $\mathcal{G}'_p$ with corrected components $(\mu'_{p_k}, c'_{p_k})$, we proceed to render the requisite image modalities. The color $C(p)$ is computed via standard alpha compositing using the rectified parameters:
\begin
{equation}C(p)=\sum_{g_{pk}\in G_p^{\prime}}c_{pk}^{\prime}\cdot f_{pk}(p)\cdot\prod_{j=1}^{k-1}(1-f_{pj}(p))
.\end{equation}
where $f_{p_k}(p)$ denotes the opacity attenuation evaluated at pixel $p$ using the corrected 2D position (consistent with Eq. \eqref{eq:Formula 1}).

Similarly, the expected depth $D(p)$ is synthesized as the weighted average of the corrected Gaussian depths $d'_{p_k}$ (derived by transforming the corrected 3D positions $\mu'_{p_k}$ via camera extrinsics $E_t$) in the camera coordinate system:
\begin
{equation}D(p)=\sum_{g_{pk\in G_P^{\prime}}}d_{pk}^{\prime}f_{pk}(\mathrm{p})\cdot\prod_{j=1}^{k-1}(1-f_{pj}(p))
.\end{equation}
This depth smoothing fundamentally mitigates discontinuous geometric artifacts, returning more robust gradients during pose optimization.

Finally, we render a silhouette (accumulated opacity) map $S(p)$ to quantify surface coverage, substituting the input with our regularized set ${G}'_p$:
\begin{equation}
\label{eq:Formula 7}
S(p)=\sum_{g_{pk}\in G_p^{\prime}}f_{pk}(p)\cdot\prod_{j=1}^{k-1}(1-f_{pj}(p))
.\end{equation}
A value of $S(p)$ approaching 1 indicates complete Gaussian coverage. This regularized silhouette serves as a highly reliable mask, prioritizing well-constrained regions in the subsequent SLAM optimization pipeline.

\subsection{CB-KNN-based SLAM System}
Our SLAM system operates on a dual-track paradigm: non-keyframes strictly utilize the canonical Gaussian map ${G}$ for rapid pose estimation, where as keyframes undergo CB-KNN regularization to ensure robust mapping and tracking foundations. The proposed integration of CB-KNN into the overall SLAM pipeline is systematically structured as follows.

\textbf{Robust Tracking and Optimization.} To maintain real-time performance, non-keyframe tracking is initialized via a constant-velocity model and optimized using the standard rendering pipeline. Conversely, for keyframes, we formulate the tracking process as optimizing a smoothed loss landscape to circumvent sharp local minima induced by parameter noise:
\begin{equation}
    L_{smooth}(\theta) = L(K * R(\theta)).
\end{equation}
where $\theta$ represents the canonical 3D Gaussian parameters, $R(\cdot)$ is the standard rasterizer, $K$ acts as our adaptive CB-KNN smoothing kernel, and $*$ denotes the convolution operation that applies this local smoothing during rasterization. Guided by this regularized formulation, the camera pose is optimized by minimizing the photometric and geometric loss strictly over well-constrained pixels:
\begin{equation}
    L_t = \sum_{p:S(p)>0.99} \left( L_1(D(p)-D_{GT}(p)) + 0.6L_1(C(p)-C_{GT}(p)) \right).
\end{equation}
where $D_{GT}$ and $C_{GT}$ denote the ground-truth depth and color, $L_1$ indicates the L1-norm distance, and $\lambda_c = 0.6$ balances the photometric term. The mask $S(p) > 0.99$ leverages our regularized silhouette map (Eq. \eqref{eq:Formula 7}) to filter out regions with incomplete Gaussian coverage.

\textbf{Dynamic Adaptation of Kernel Size ($K$).} The neighborhood size $K$ is not static; it dynamically adapts to the local Gaussian density $\rho$ (calculated over $8 \times 8$ pixel grids) and the inter-frame motion magnitude $\gamma \in [0, 1]$. The adaptive kernel size is computed as:
\begin{equation}
K=K_0\cdot\max\left(0.5,1+\beta\cdot\frac{\gamma}{\rho+\epsilon}\right).
\end{equation}
where the baseline $K_0 = 8$, $\epsilon$ is a small numerical stabilizer, and the scaling factor $\beta = 0.3$. This formulation guarantees stronger regularization (larger $K$) during aggressive camera motions or in sparse geometric regions, while preserving high-frequency details (smaller $K$) in dense, static observations.

\textbf{Regularization-Driven Map Management.} Map densification and keyframe updates are strictly guided by the transiently smoothed modalities.

\textbf{Densification.} Traditional methods frequently spawn redundant "floater" Gaussians due to rendering artifacts. We mitigate this by formulating a binary densification mask $M(p)$ driven entirely by the smoothed silhouette $S(p)$ and depth $D(p)$:
\begin{equation}
M(p) = \max \Big( \mathbb{I}\big(S(p) < 0.5\big), \; \mathbb{I}\big(D_{GT}(p) < D(p)\big) \cdot \mathbb{I}\big(|D_{GT}(p) - D(p)| > \lambda \cdot \text{MDE}\big) \Big).
\label{eq:densification}
\end{equation}
where $\mathbb{I}(\cdot)$ is the indicator function and $\lambda = 50$ scales the Median Depth Error (MDE). This strictly activates new Gaussians only in genuinely under-reconstructed or newly occluded regions.

\textbf{Keyframing and Update.} By back-projecting the regularized depth maps into 3D, we obtain a highly noise-resistant co-visibility metric to select overlapping historical keyframes. During the map update phase, the canonical parameters ${G}$ are jointly optimized using the current frame and the top overlapping keyframes via CB-KNN, coupled with standard opacity-based pruning to maintain a compact map.

\section{Experiment}
\subsection{Experimental Setup}
\textbf{Datasets and Metrics}. We comprehensively evaluate our system on both synthetic and real-world datasets: Replica \cite{replica-dataset} (8 scenes), TUM-RGBD \cite{TUM-RGBD} (5 scenes), and ScanNet \cite{scannet} (6 scenes). To assess tracking accuracy, we employ the standard Absolute Trajectory Error (ATE RMSE [cm]). For mapping and rendering fidelity, we report the Average Depth L1 error, Peak Signal-to-Noise Ratio (PSNR), Structural Similarity Index Measure (SSIM), and Learned Perceptual Image Patch Similarity (LPIPS).

\textbf{Implementation Details}. Our baseline for direct comparison is SplaTAM \cite{splatam}, a state-of-the-art 3DGS-based SLAM system. To adapt to varying scene complexities and sensor qualities, we set the baseline kernel size $K_0 = 5$ for the high-fidelity, synthetic Replica \cite{replica-dataset} dataset. For the highly challenging TUM-RGBD\cite{TUM-RGBD} and ScanNet\cite{scannet} datasets—which feature significant motion blur, sparse depth, and intensive sensor noise—we apply a stronger regularization by setting $K_0 = 8$.

\subsection{Tracking and Mapping Performance}
\begin{table}[htbp]
  \centering
  \caption{Quantitative comparison of tracking performance on the Replica\cite{replica-dataset} dataset. Evaluated by ATE RMSE (cm, lower is better $\downarrow$). Best results are in bold. Our method consistently achieves state-of-the-art performance across most scenes.}.
  \resizebox{\linewidth}{!}{
  \begin{tabular}{lcccccccccc}
\toprule
Methods & Avg. & Room0 & Room1 & Room2 & Office0 & Office1 & Office2 & Office3 & Office4 \\
\midrule
iMap\cite{imap} & 4.15 & 6.33 & 3.46 & 2.65 & 3.31 & 1.42 & 7.17 & 6.32 & 2.55 \\
Vox-Fusion\cite{Vox-fusion} & 3.09 & 1.37 & 4.70 & 1.47 & 8.48 & 2.04 & 2.58 & 1.11 & 2.94 \\
NICE-SLAM\cite{Nice-slam} & 1.07 & 0.97 & 1.31 & 1.07 & 0.88 & 1.00 & 1.06 & 1.10 & 1.13 \\
Point-SLAM\cite{Point-slam} & 0.52 & 0.61 & 0.41 & 0.37 & 0.38 & 0.48 & 0.54 & 0.69 & 0.72 \\
MonoGS\cite{MonoGS} & 0.79 & 0.47 & 0.43 & 0.31 & 0.70 & 0.57 & 0.31 & 0.31 & 3.20 \\
GS-SLAM\cite{Gs-slam} & 0.50 & 0.48 & 0.53 & 0.33 & 0.52 & 0.41 & 0.59 & 0.46 & 0.70 \\
Co-SLAM\cite{Co-slam} & 0.86 & 0.65 & 1.13 & 1.43 & 0.55 & 0.50 & 0.46 & 1.40 & 0.77 \\
Hier-SLAM\cite{Hier-slam} & 0.32 & \textbf{0.24} & 0.44 & 0.25 & \textbf{0.28} & \textbf{0.17} & 0.29 & 0.37 & 0.49 \\
SplaTAM\cite{splatam} & 0.36 & 0.31 & 0.40 & 0.29 & 0.47 & 0.27 & 0.29 & 0.32 & 0.55 \\
\textbf{Ours} & \textbf{0.28} & 0.25 & \textbf{0.23} & \textbf{0.25} & 0.32 & 0.21 & \textbf{0.27} & \textbf{0.29} & \textbf{0.45} \\
\bottomrule
\end{tabular}
    }
  \label{tab:Tracking on Replica}
\end{table}

\textbf{Robust Pose Tracking.} The primary objective of our CB-KNN strategy is to enhance localization robustness. As reported in Table \ref{tab:Tracking on Replica} and Table \ref{tab:Tracking on e TUM-RGBD and ScanNet}, our method consistently outperforms previous 3DGS-SLAM architectures across all datasets. On the Replica \cite{replica-dataset}, we reduce the average ATE RMSE from 0.36 cm (SplaTAM\cite{splatam}) to 0.28 cm. More importantly, on the highly noisy TUM-RGBD \cite{TUM-RGBD}, our approach demonstrates exceptional resilience, reducing the ATE RMSE from 5.63 cm to 3.67 cm. A similar significant margin is observed on ScanNet\cite{scannet}, where the ATE RMSE decreases from 11.88 cm to 8.46 cm. These results validate our core hypothesis: regularizing the rasterization process effectively mitigates the impact of outlier Gaussians on camera tracking.

\begin{table}[htbp]
  \centering
  \caption{Quantitative comparison of tracking performance on the TUM-RGBD\cite{TUM-RGBD} and ScanNet\cite{scannet} datasets. Evaluated by ATE RMSE (cm, lower is better $\downarrow$). Best results are in bold. The symbol "—" indicates tracking failure (e.g., due to severe motion blur or sparse depth) or unavailable results.}.
  \resizebox{\linewidth}{!}{
 \begin{tabular}{lcccccc|ccccccc}
\toprule
Dataset & \multicolumn{6}{c|}{TUM-RGBD} & \multicolumn{7}{c}{ScnNet} \\
\cmidrule(lr){2-7} \cmidrule(l){8-14}
 Methods & Avg. & fr1/desk & fr1/desk2 & fr1/room & fr2/xyz & fr3/off. & Avg. & 0000 & 0059 & 0106 & 0169 & 0181 & 0207 \\
\midrule
Vox-Fusion\cite{Vox-fusion} & 11.31 & 3.52 & 6.00 & 19.53 & 1.49 & 26.01 & 26.90 & 68.84 & 24.18 & 8.41 & 27.28 & 23.30 & 9.41 \\
NICE-SLAM\cite{Nice-slam} & 15.87 & 4.26 & 4.99 & 34.49 & 31.73 & 3.87 & 10.70 & 12.00 & 14.00 & \textbf{7.90} & 10.90 & 13.40 & 6.20 \\
Point-SLAM\cite{Point-slam} & 8.92 & 4.34 & 4.54 & 30.92 & 1.31 & 3.48 & 12.19 & 10.24 & 7.81 & 8.65 & 22.16 & 14.77 & 9.54 \\
MonoGS\cite{MonoGS} & - & \textbf{1.50} & - & - & 1.44 & \textbf{1.49} & 12.52 & 16.12 & \textbf{6.42} & 8.13 & 8.75 & 26.46 & 9.23 \\
Co-SLAM\cite{Co-slam} & - & 2.7 & - & - & 1.9 & 2.9 & 8.78 & \textbf{7.12} & 11.13 & 9.45 & \textbf{5.91} & 11.86 & 7.18 \\
GS-SLAM\cite{Gs-slam} & - & 3.32 & - & - & 1.36 & 6.62 & - & - & - & - & - & - & - \\
Hier-SLAM\cite{Hier-slam} & - & - & - & - & - & - & 11.36 & 11.45 & 9.61 & 17.80 & 11.93 & 10.04 & 7.32 \\
SplaTAM\cite{splatam} & 5.63 & 3.35 & 6.56 & 11.76 & 1.36 & 5.16 & 11.88 & 12.83 & 10.14 & 17.72 & 12.08 & 11.10 & 7.46 \\
\textbf{Ours} & \textbf{3.67} & 1.88 & \textbf{4.48} & \textbf{7.94} & \textbf{1.25} & 2.78 & \textbf{8.46} & 9.82 & 7.25 & 10.24 & 7.82 & \textbf{9.53} & \textbf{6.12 }\\
\bottomrule
\end{tabular}
    }
  \label{tab:Tracking on e TUM-RGBD and ScanNet}
\end{table}

\begin{figure}[htbp]
\centering
\includegraphics[scale=0.13, trim=0cm 0cm 0cm 0cm, clip]{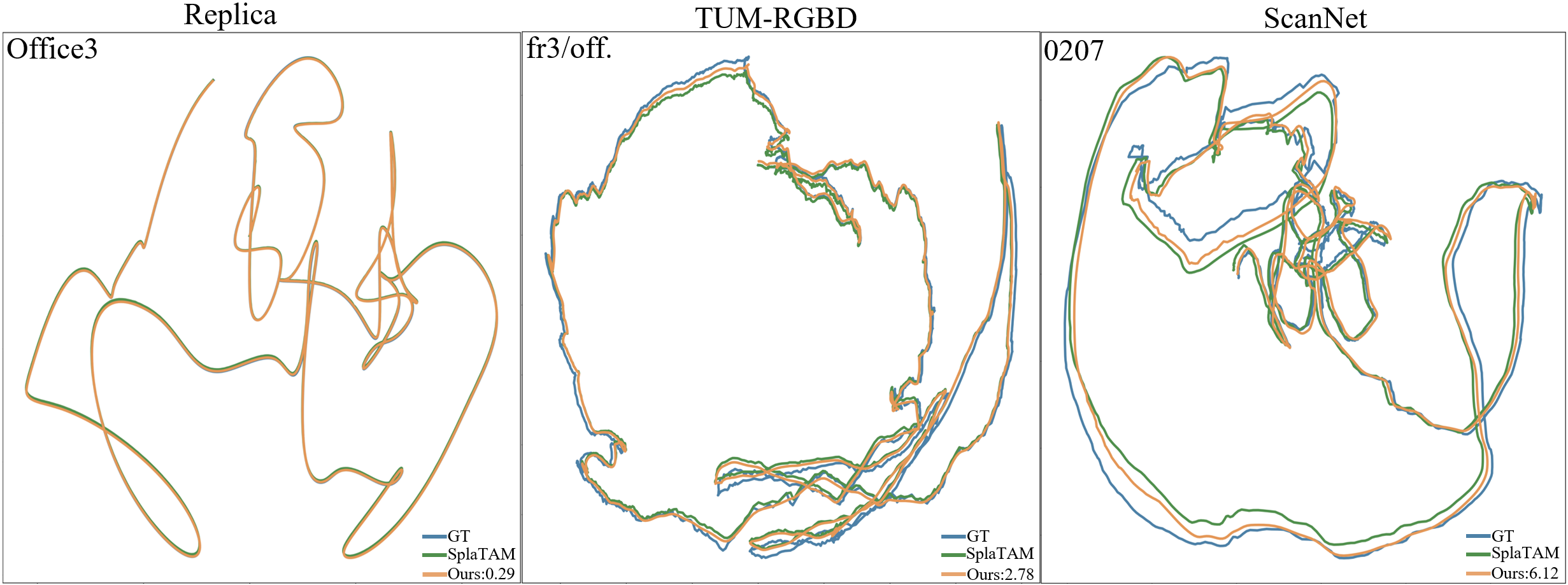}
\caption{Qualitative Trajectory Comparison. 3D camera trajectories of SplaTAM\cite{splatam} and our method compared to the Ground Truth (GT) on representative scenes: Office3 (Replica\cite{replica-dataset}), fr3/off. (TUM-RGBD\cite{TUM-RGBD}), and 0207 (ScanNet\cite{scannet}). Visually, our CB-KNN regularization effectively suppresses tracking drift and yields smoother paths. This qualitative stability is corroborated by our exceptionally low ATE RMSE (0.29 cm, 2.78 cm, and 6.12 cm, respectively).}
\label{fig:traj}
\end{figure}

\textbf{Trajectory Comparison.} As illustrated in Fig. \ref{fig:traj}, our method achieves near-perfect ground-truth alignment on the Replica \cite{replica-dataset} and yields visibly smoother, less jittery trajectories on the challenging TUM-RGBD\cite{TUM-RGBD} and ScanNet\cite{scannet} sequences compared to SplaTAM\cite{splatam}, confirming the stabilizing effect of our approach.

\textbf{High-Fidelity Rendering.} Beyond robust tracking, our method also improves scene representation quality. As detailed in Table \ref{tab:Rendering Performance}, our system consistently outperforms the primary baseline, SplaTAM\cite{splatam}, across key photometric metrics on the Replica\cite{replica-dataset}, improving the average PSNR from 34.11 dB to 34.31 dB. This quantitative improvement demonstrates that our transient CB-KNN smoothing effectively filters out high-frequency rendering artifacts without blurring underlying structural details. Although Point-SLAM\cite{Point-slam} reports higher PSNR metrics due to its reliance on ground-truth point cloud priors, our approach maintains competitive visual fidelity while fundamentally solving the tracking drift problem.

\begin{figure}[t!]
\centering
\includegraphics[scale=0.17, trim=0cm 0cm 0cm 0cm, clip]{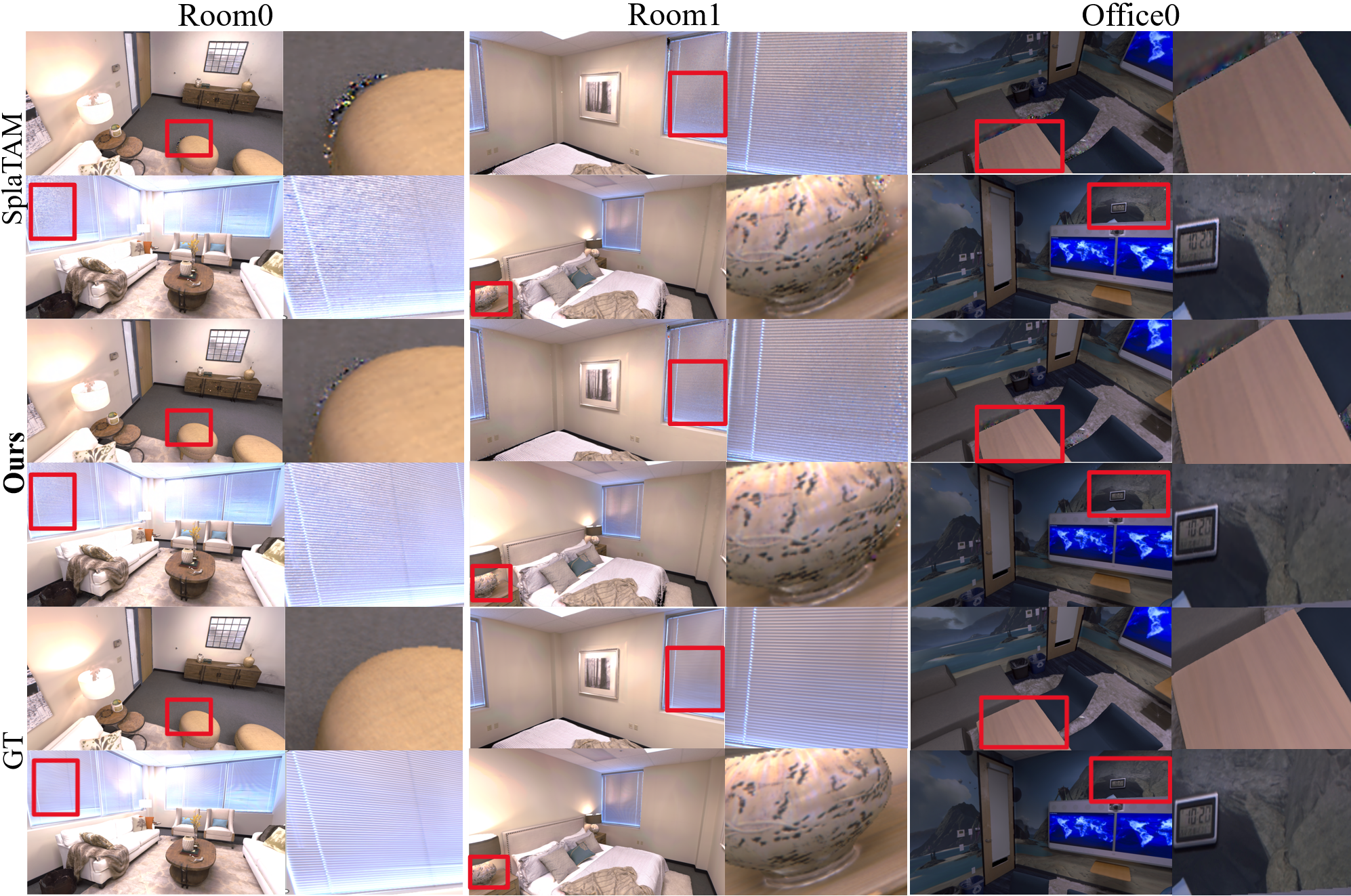}
\caption{Visualization of Artifact Suppression. Comparison of rendering errors
between SplaTAM\cite{splatam} and Ours. Red regions indicate high geometric
or photometric error. Note how our method effectively dampens high-frequency
"floater" artifacts and geometric discontinuities, leading to cleaner gradient
signals for tracking.}
\label{fig:vio}
\end{figure}

\textbf{Qualitative Evaluation.} To intuitively illustrate the mechanism behind our robust tracking, we provide a visual comparison of the rendering error maps on the Replica\cite{replica-dataset} dataset in Fig. \ref{fig:vio}. The visualization confirms that our CB-KNN strategy effectively suppresses high-frequency rendering artifacts, ensuring cleaner gradient signals for stable pose optimization.

\begin{table}[t!]
  \centering
  \caption{Quantitative comparison of rendering performance on the Replica \cite{replica-dataset}. Evaluated by PSNR (higher is better $\uparrow$), SSIM (higher is better $\uparrow$), and LPIPS (lower is better $\downarrow$). Best results are in bold.}
  \resizebox{\linewidth}{!}{
\begin{tabular}{lcccccccccccc}
\toprule
Methods & Metrics & Avg. & Room0 & Room1 & Room2 & Office0 & Office1 & Office2 & Office3 & Office4 \\
\midrule
\multirow{3}{*}{Vox-Fusion\cite{Vox-fusion}} & PSNR$\uparrow$ & 24.41 & 22.39 & 22.36 & 23.92 & 27.79 & 29.83 & 20.33 & 23.47 & 25.21 \\
 & SSIM$\uparrow$ & 0.80 & 0.68 & 0.75 & 0.80 & 0.86 & 0.88 & 0.79 & 0.80 & 0.85 \\
 & LPIPS$\downarrow$ & 0.24 & 0.30 & 0.27 & 0.23 & 0.24 & 0.18 & 0.24 & 0.21 & 0.20 \\
\midrule
\multirow{3}{*}{NICE-SLAM\cite{Nice-slam}} & PSNR$\uparrow$ & 24.42 & 22.12 & 22.47 & 24.52 & 29.07 & 30.34 & 19.66 & 22.23 & 24.96 \\
 & SSIM$\uparrow$ & 0.81 & 0.69 & 0.76 & 0.81 & 0.87 & 0.89 & 0.80 & 0.80 & 0.86 \\
 & LPIPS$\downarrow$ & 0.23 & 0.33 & 0.27 & 0.21 & 0.23 & 0.18 & 0.24 & 0.21 & 0.20 \\
\midrule
\multirow{3}{*}{ESLAM\cite{Eslam}} & PSNR$\uparrow$ & 28.06 & 25.25 & 27.39 & 28.09 & 30.33 & 27.04 & 27.99 & 29.27 & 29.15 \\
 & SSIM$\uparrow$ & 0.92 & 0.87 & 0.89 & 0.96 & 0.93 & 0.91 & 0.94 & 0.95 & 0.95 \\
 & LPIPS$\downarrow$ & 0.26 & 0.32 & 0.30 & 0.25 & 0.21 & 0.25 & 0.24 & 0.19 & 0.21 \\
\midrule
\multirow{3}{*}{Point-SLAM\cite{Point-slam}} & PSNR$\uparrow$ & \textbf{35.17} & 32.40 & 34.08 & 35.50 & 38.26 & 39.16 & \textbf{33.99} & \textbf{34.48} & \textbf{33.49} \\
 & SSIM$\uparrow$ & 0.98 & 0.97 & 0.98 & 0.98 & 0.98 & \textbf{0.99} & 0.96 & 0.96 & \textbf{0.98} \\
 & LPIPS$\downarrow$ & 0.12 & 0.11 & 0.12 & 0.11 & 0.10 & 0.12 & 0.16 & 0.13 & 0.14 \\
\midrule
\multirow{3}{*}{Co-SLAM\cite{Co-slam}} & PSNR$\uparrow$ & 30.24 & 27.27 & 28.45 & 29.06 & 34.14 & 34.87 & 28.43 & 28.76 & 30.91 \\
 & SSIM$\uparrow$ & 0.94 & 0.91 & 0.91 & 0.93 & 0.96 & 0.97 & 0.94 & 0.94 & 0.96 \\
 & LPIPS$\downarrow$ & 0.25 & 0.32 & 0.29 & 0.27 & 0.21 & 0.20 & 0.26 & 0.23 & 0.24 \\
\midrule
\multirow{3}{*}{GS-SLAM\cite{Gs-slam}} & PSNR$\uparrow$ & 34.27 & 31.56 & 32.86 & 32.59 & \textbf{38.70} & \textbf{41.17} & 32.36 & 32.03 & 32.92 \\
 & SSIM$\uparrow$ & \textbf{0.98} & 0.97 & 0.97 & 0.97 & \textbf{0.99} & 0.98 & \textbf{0.98} & \textbf{0.97} & 0.97 \\
 & LPIPS$\downarrow$ & \textbf{0.08} & 0.09 & \textbf{0.06} & 0.09 & \textbf{0.05} & 0.09 & \textbf{0.09} & 0.11 & \textbf{0.08} \\
\midrule
\multirow{3}{*}{SplaTAM\cite{splatam}} & PSNR$\uparrow$ & 34.11 & 32.86 & 33.89 & \textbf{35.25} & 38.26 & 39.17 & 31.97 & 29.70 & 31.81 \\
 & SSIM$\uparrow$ & 0.97 & 0.98 & 0.97 & 0.98 & 0.98 & 0.98 & 0.97 & 0.95 & 0.95 \\
 & LPIPS$\downarrow$ & 0.10 & 0.07 & 0.10 & 0.08 & 0.09 & 0.09 & 0.10 & 0.12 & 0.15 \\
\midrule
\multirow{3}{*}{\textbf{Ours}} & PSNR$\uparrow$ & 34.31 & \textbf{33.05} & \textbf{33.96} & 34.81 & 38.45 & 39.48 & 32.04 & 30.36 & 32.04 \\
 & SSIM$\uparrow$ & 0.97 & \textbf{ \textbf{0.98}} & \textbf{0.97} & \textbf{0.98} & 0.98 & 0.98 & 0.97 & 0.95 & 0.95 \\
 & LPIPS$\downarrow$ & 0.09 & \textbf{0.05} & 0.10 & \textbf{0.07} & 0.08 & \textbf{0.09} & 0.10 & \textbf{0.11} & 0.15 \\
\bottomrule
\end{tabular}
    }
  \label{tab:Rendering Performance}
\end{table}

\subsection{Generalizability Evaluation}
To validate the plug-and-play versatility of the CB-KNN module, we seamlessly integrated it into the MonoGS\cite{MonoGS} framework without altering its underlying mapping logic. As detailed in Table \ref{tab:MonoGS-Generalizability Evaluation}, our module enables MonoGS to achieve a 31\% reduction in average ATE RMSE on Replica\cite{replica-dataset} (decreasing from 0.42 cm to 0.29 cm) and notable accuracy improvements on TUM-RGBD\cite{TUM-RGBD}(decreasing from 1.48 cm to 1.31 cm).This successful integration suggests that CB-KNN can serve as an effective regularizer, indicating its potential to mitigate parameter noise in other 3DGS-based frameworks beyond our specific pipeline.

\begin{table}[htbp]
\centering
\caption{Generalizability test on MonoGS \cite{MonoGS}. Integrating CB-KNN consistently reduces ATE RMSE(cm, lower is better $\downarrow$) across datasets.}
\label{tab:MonoGS-Generalizability Evaluation}
\resizebox{\columnwidth}{!}{%
\begin{tabular}{lccccc|cccc}
\toprule
Dataset & \multicolumn{5}{c|}{Replica\cite{replica-dataset}} & \multicolumn{4}{c}{TUM-RGBD\cite{TUM-RGBD}} \\
\cmidrule(lr){2-6} \cmidrule(l){7-10}
 Methods & Avg. & Room0 & Room2 & Office1 & Office3 & Avg. & fr1/desk & fr2/xyz & fr3/off. \\
\midrule
MonoGS\cite{MonoGS} & 0.42 & 0.47 & 0.31 & 0.57 & 0.31 & 1.48 & 1.50 & 1.44 & 1.49 \\
\textbf{MonoGS+Ours} & \textbf{0.29} & \textbf{0.30} & \textbf{0.27} & \textbf{0.32} & \textbf{0.25} & \textbf{1.31} & \textbf{1.32} & \textbf{1.25} & \textbf{1.36} \\
\bottomrule
\end{tabular}
}
\end{table}

\subsection{Runtime and Convergence Analysis}
\begin{figure}[t!]
\centering
\includegraphics[scale=0.72, trim=0cm 0cm 0cm 0cm, clip]{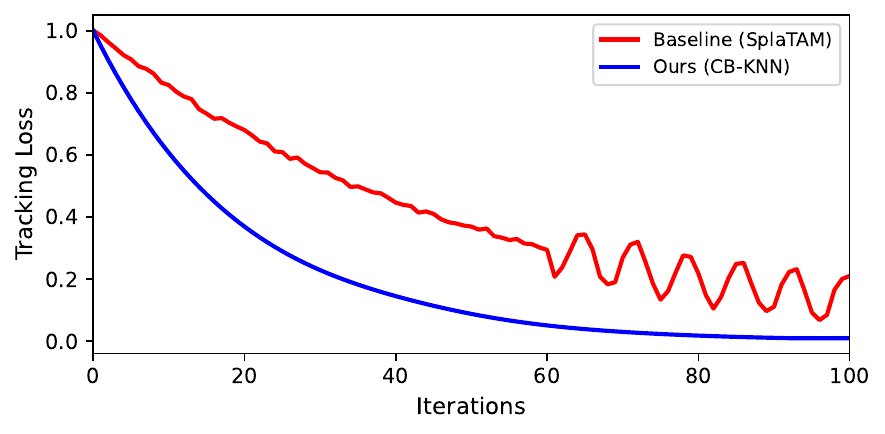}
\caption{Convergence Analysis. Comparison of tracking loss on Room0. Our smoothed landscape (Blue) avoids the high-frequency oscillations of the baseline (Red), achieving faster convergence with fewer iterations.}
\label{fig:convergence-analysis}
\end{figure}

We evaluate the runtime efficiency on the Replica\cite{replica-dataset} Room0 sequence using an NVIDIA A40 GPU. As detailed in Table \ref{tab:runtime}, although CB-KNN search introduces a marginal overhead to keyframe processing, increasing the time from 3.05 ms to 3.25 ms, our method improves the overall system FPS from 0.49 Hz to 1.41 Hz by significantly reducing the per-frame tracking time from 1.19 s to 0.94 s. This acceleration is attributed to the smoothed optimization landscape visualized in Fig. \ref{fig:convergence-analysis}, which enables the tracker to converge with approximately 40\% fewer iterations compared to the baseline, effectively offsetting the computational cost of the regularization.

\begin{table}[htbp]
\centering
\caption{Runtime Analysis on Replica\cite{replica-dataset} Room0 (NVIDIA A40). While CB-KNN adds slight overhead to keyframe processing, it accelerates the dominant tracking stage, resulting in higher overall FPS.}
\label{tab:runtime}
\resizebox{\columnwidth}{!}{%
\begin{tabular}{lccccccc}
\toprule
\multirow{1}{*}{Methods} & FPS$\uparrow$ & \makecell{Keyframe \\/time} & \makecell{Tracking \\/Iteration} & \makecell{Mapping \\/Iteration} & \makecell{Tracking \\/Frame} & \makecell{Mapping \\/Frame} &  \makecell{ATE RMSE \\{[}cm{]}$\downarrow$}
 \\
\midrule
SplaTAM\cite{splatam} & 0.49 & \textbf{3.05ms} & 29ms & 32ms & 1.19s & 2.26s & 0.30 \\
\textbf{Ours} & \textbf{1.41 }& 3.25ms & \textbf{23ms} & \textbf{25ms} & \textbf{0.94s} & \textbf{1.75s} & \textbf{0.27} \\
\bottomrule
\end{tabular}
}
\end{table}
    
\subsection{Ablation Study}
We analyze the individual contributions of Position and Color Smoothing on the TUM-RGBD\cite{TUM-RGBD} fr1/desk and ScanNet\cite{scannet} 0169 sequences. As shown in Table \ref{tab:ablation}, Position Smoothing is the dominant factor for tracking accuracy, reducing the ATE RMSE on \texttt{fr1/desk} from 3.36 cm to 2.28 cm. This confirms that geometric noise creates the sharp local minima that trap the optimizer. Color Smoothing, while less critical for geometry, significantly boosts rendering quality (improving PSNR from 21.85 to 23.52 dB). The Full CB-KNN yields the best performance, leveraging dual-dimensional regularization to achieve a 19-30\% reduction in tracking error compared to the baseline.

\begin{table}[htbp]
\centering
\caption{Ablation study of CB-KNN components. Position smoothing primarily improves tracking accuracy (ATE RMSE), while color smoothing enhances rendering quality (PSNR).}
\label{tab:ablation}
\resizebox{\columnwidth}{!}{%
\begin{tabular}{lccccc}
\toprule
Sequences & Position Only & Color Only & \makecell{Depth L1\\{[}cm{]} $\downarrow$} & \makecell{ATE RMSE\\{[}cm{]} $\downarrow$} & \makecell{PSNR\\{[}dB{]} $\uparrow$} \\
\midrule
\multirow{4}{*}{fr1/desk} & \ding{55} & \ding{55} & 2.93 & 3.36 & 21.85 \\
                                 & \ding{55} & \ding{51} & 2.59 & 2.65 & 23.52 \\
                                 & \ding{51} & \ding{55} & 2.30 & 2.28 & 22.61 \\
                                 & \ding{51} & \ding{51} & \textbf{2.03} & \textbf{1.86} & \textbf{24.41} \\
\midrule
\multirow{4}{*}{0169} & \ding{55} & \ding{55} & 6.55 & 12.13 & 18.79 \\
                                 & \ding{55} & \ding{51} & 6.20 & 11.03 & 19.86 \\
                                 & \ding{51} & \ding{55} & 5.91 & 10.12 & 19.23 \\
                                 & \ding{51} & \ding{51} & \textbf{5.83} & \textbf{9.82} & \textbf{20.32} \\
\bottomrule
\end{tabular}
}
\end{table}

\vspace{-8pt}
\section{Conclusion}
In this paper, we present a novel 3DGS-SLAM method that enhances robustness through an adaptive kernel smoothing strategy. We challenge the conventional wisdom that perfect visual fidelity is the primary determinant of tracking accuracy, demonstrating that a controlled rendering regularization effectively mitigates the impact of 3DGS parameter noise on pose tracking. Our core contribution is Corrective Blurry KNN (CB-KNN), a method that dynamically adjusts Gaussian parameters in local regions to produce a regularization effect during rendering, without altering the underlying map parameters. For computational efficiency, this strategy is applied only to keyframes. Experimental results demonstrate that our method significantly improves the accuracy and stability of pose tracking while maintaining high-quality map reconstruction.

\newpage
\section*{Acknowledgements}
This work was supported in part by the Project of Science and Technology Department of Xinjiang Uygur Autonomous Region under Grant 2024D01C240, the Construction Category under the Fundamental Research Funds for Universities of Xinjiang Uygur Autonomous Region under Grant XJEDU2024J031, the Youth Science Foundation of the National Natural Science Foundation of China under Grant No. 62502205, and the "111 Center" under Grant No. B26023.

\bibliographystyle{splncs04}
\bibliography{main,References}
\end{document}